\documentclass{article}

\PassOptionsToPackage{numbers,compress}{natbib}

\usepackage[dblblindworkshop, final]{neurips_2026}

\makeatletter
\renewcommand{\@noticestring}{AI Agents for Discovery in the Wild (AID-Wild), Workshop at ACM CAIS 2026.}
\makeatother

\usepackage[utf8]{inputenc}
\usepackage[T1]{fontenc}
\usepackage{hyperref}
\usepackage{url}
\usepackage{booktabs}
\usepackage{amsfonts}
\usepackage{nicefrac}
\usepackage{microtype}
\usepackage{xcolor}

\title{Interpretable Early Termination of Web Navigation Agents via Closed Sequential Pattern Mining}
\workshoptitle{AI Agents for Discovery in the Wild}

\author{%
  Sergio Talavera \\
  Department of Computer Engineering \\
  San Jose State University \\
  San Jose, CA 95192 \\
  \texttt{sergio.talavera@sjsu.edu} \\
  \And
  Magdalini Eirinaki \\
  Department of Computer Engineering \\
  San Jose State University \\
  San Jose, CA 95192 \\
  \texttt{magdalini.eirinaki@sjsu.edu}
}

\begin{document}

\maketitle

\begin{abstract}
Web navigation agents built on open-weight language models fail often and fail late, with most of the tokens spent after the trajectory has become unrecoverable. Recent work has shown that early termination is a viable cost-reduction strategy, but existing approaches produce opaque termination decisions that do not support diagnosis or human oversight. In this work we apply closed sequential pattern mining (BIDE+) to $K$-step prefixes of agent execution traces to produce a small, human-reviewable failure-signature library that organizes into a four-category failure taxonomy with associated repair recommendations. We use a multi-variant evaluation protocol that explicitly removes contamination artifacts (zero-step crashes and trivially separable single-step successes) so that reported detection accuracy reflects behavioral discrimination rather than trace-length distributions. We evaluate our methodology on 1{,}544 MiniWoB++ traces collected from three open-weight language models, achieving the best F1 at $K{=}3$ on the Exclude-errors configuration across eight methods (0.717), recovering 88\% of test-set failures within three steps, and supporting an early-termination policy that saves 26.3\% of tokens at 92\% precision. %We also report the failure modes of the approach: per-model F1 spans 0.526--0.913, transfer to WebArena is model-dependent, and a bidirectional LSTM exceeds BIDE on F1 at $K{=}10$ when the evaluation includes timeouts.
\end{abstract}

\section{Introduction}
\label{sec:intro}

Large language models have made it practical to build autonomous agents that execute multi-step tasks on the web. Architectures such as ReAct~\cite{yao2023react} interleave a natural-language reasoning trace with each action, producing structured execution traces. Despite progress, current agents fail often: Zhou et al.~\cite{zhou2024webarena} report 14.4\% task success on WebArena for GPT-4 against 78.24\% for human evaluators, and open-weight models in the 3--7B parameter range, the setting evaluated here, perform substantially worse.

Most of the cost of a failing run is incurred late in the trace. Xiao et al.~\cite{xiao2025trajectory} observe that approximately 99\% of token consumption in agent workflows comes from accumulated input tokens replayed at each step, so per-step cost grows as the trace proceeds. When an agent commits to an unrecoverable trajectory early, most of its tokens are spent after failure has become inevitable.

Two recent systems address this cost directly. AgentDiet~\cite{xiao2025trajectory} compresses the trajectory during execution; EET~\cite{guo2026eet} retrieves structured experience to terminate software-engineering agents when patch generation looks unlikely to succeed. Both establish early intervention as a viable cost-reduction strategy, but neither produces an interpretable account of \emph{why} a particular run is being terminated; the decision reduces to a scalar score or retrieval similarity. This is insufficient for deployment settings where operators audit behavior, attribute failures to architectural defects, or reuse the signal for diagnostic workflows.

In this work we try to address this by proposing an explainable methodology that enables early stops. We mine closed sequential patterns from $K$-step prefixes of labeled agent traces. The result is a small library of failure-predictive subsequences (32 patterns, organized into four behavioral categories) that supports both an early-termination policy and a diagnostic readout. Each termination decision is justified by the specific patterns that matched, their positions in the trace, and the failure category they encode.

This paper makes four contributions. First, a pattern-mining approach to early failure detection for LLM-based web navigation agents, evaluated against seven baselines. Second, a multi-level symbolization scheme that converts heterogeneous reasoning--action--observation traces into discrete sequences with controllable abstraction. Third, a multi-variant evaluation protocol that progressively removes contamination categories (zero-step error traces, single-step successes), isolating behavioral discrimination from trace-length artifacts. Fourth, a four-category failure taxonomy with repair recommendations derived from the mined library, and an operating-point analysis showing 26.3\% token savings at 92\% precision in the deployment configuration.

\section{Methodology}
\label{sec:method}

The system has two phases. In the offline phase, labeled training traces are symbolized, truncated to $K$-step prefixes, mined for closed sequential patterns, filtered by failure precision, and organized into a taxonomy. In the online phase, each step of a live trace is symbolized incrementally and matched against the library; a coverage score above a deployment threshold triggers termination with an interpretable justification.

Each trace step is encoded along four dimensions: the action type (\textsc{click}, \textsc{type}, \textsc{navigate}, etc.), the selector strategy used to address the target element, the outcome observed in the environment, and a reasoning intent extracted from the agent's chain-of-thought text by keyword and regex matching against a fixed vocabulary (e.g., \textsc{verify}, \textsc{retry}, \textsc{stuck}). Traces are labeled by the environment's per-episode task-success signal: successes are episodes the environment marks as completed; failures are episodes that end without completion within the 30-step budget. Three abstraction levels (fine, medium, coarse) trade granularity for robustness to non-determinism from LLM sampling and DOM rendering. The reported experiments use the medium level of abstraction (e.g., \texttt{CLICK\_BID\_SUCCESS} or \texttt{CLICK\_BID\_SUCCESS\_\_R\_VERIFY}).

Each symbolic trace is truncated to its first $K$ symbols. We apply BIDE+~\cite{wanghanli2007bideplus} via SPMF~\cite{fournierviger2016spmf} at 5\% minimum support. BIDE mines only closed patterns directly without candidate maintenance, where a pattern is closed if no longer pattern occurs in exactly the same set of traces; this removes redundant sub-patterns while preserving the support information. At 5\% support, BIDE produces 33 closed patterns, a library small enough to inspect and organize by hand. Each pattern is scored by its precision on the failure class (fraction of training prefixes containing it that are labeled failures) and patterns scoring below 0.50 are dropped. At inference, library patterns are matched against the live prefix as ordered subsequences (not necessarily contiguous). Matched precision scores aggregate into a coverage score and when coverage exceeds a tuned threshold the run is terminated. The system returns the matched patterns, their positions, and their failure category alongside the termination decision.

Manual inspection of the mined library yielded four failure categories: a) recovery failures contain unparseable-action symbols (e.g.\ \texttt{UNKNOWN\_NONE\_SUCCESS}), reflecting loss of coherence and fallback to repetitive actions; b) validation failures contain repeated clicks tagged with verification reasoning, reflecting re-checking loops; c) navigation failures consist of extended click sequences without other action types, reflecting interaction without task progress; and d) context failures involve excessive repetitive typing or premature stop signals, reflecting loss of task state. Each category maps to a concrete repair recommendation at the agent-architecture level.

\section{Experimental Setup}
\label{sec:setup}

We collected 1{,}544 MiniWoB++~\cite{liu2018miniwob} traces (621 success, 290 failure, 315 timeout, 318 error) across 10 medium-difficulty tasks using three open-weight backbones: Llama-3.2-3B, Qwen-2.5-7B, and Mistral-7B-Instruct-v0.3. An additional 577 WebArena~\cite{zhou2024webarena} shopping traces (193 Llama, 192 Qwen, 192 Mistral; 13 successes, 174 natural failures, 254 timeouts, 136 errors) were collected for a cross-benchmark transfer analysis. The agent uses ReAct prompting with a 30-step budget, and step-level token counts averaged 875.7 tokens per step.

An initial pilot on easy MiniWoB++ tasks surfaced two contamination artifacts: zero-step error traces (process crashes before any action) and single-step successes, both of which let length-aware classifiers achieve high F1 scores by predicting failure for any trace longer than one step. The medium-task corpus avoids the latter by design. We evaluate on three configurations of increasing difficulty: \emph{Full} (all categories), \emph{Exclude-errors} (errors removed; timeouts retained as failures), and \emph{Variant C} (errors and timeouts both removed; natural failures versus successes only). The detection comparison in Table~\ref{tab:variant-c} uses the class-balanced split of Variant C. The savings analysis in Table~\ref{tab:savings} uses the natural-prevalence Exclude-errors test set (245 traces, 121 failures and 124 successes). A step-count baseline that predicts failure proportionally to trace length serves as a control: methods that fail to exceed it are not extracting behavioral signal. 

We compare BIDE Coverage against frequency vectors with logistic regression, $n$-gram features ($n{=}1{-}3$), DeepLog~\cite{du2017deeplog}, a bidirectional LSTM~\cite{graves2005bilstm}, TaSPM~\cite{taspm2024} targeted pattern mining, process conformance checking, and the step-count control. Data is split 60/20/20 with stratified sampling at fixed seed; per-method thresholds are tuned on validation by maximizing macro-F1 over a 200-point sweep. Prefix lengths were set to $K \in \{3,5,8,10\}$.

\section{Results}
\label{sec:results}

Table~\ref{tab:variant-c} reports F1@$K$ on Variant C, which isolates the core detection problem (natural failures vs.\ multi-step successes; no timeouts). The step-count baseline achieves exactly 0.667 (the majority-class F1) at all $K$, confirming no behavioral signal is available from trace length alone. BIDE Coverage leads from $K{=}5$ through $K{=}10$ (0.683 / 0.678 / 0.678). The bidirectional LSTM, competitive when timeouts are included as failures, drops to 0.631 at $K{=}10$ on Variant C, suggesting its earlier advantage partly reflects timeout-correlated features. On Exclude-errors (timeouts retained), BIDE achieves the highest F1 at $K{=}3$ (0.717) but is overtaken by Bi-LSTM at $K{=}10$ (0.612 vs.\ 0.765)\footnote{Detailed results are omitted due to space limitations; the same applies to results in Section~\ref{sec:limitations}. Extended results will be referenced upon de-anonymization.} 

\begin{table}[t]
  \centering
  \caption{Variant C balanced (natural failures vs.\ successes). BIDE Coverage leads from $K{=}5$. Step Count is a length-only control: methods that fail to exceed it are not extracting behavioral signal.}
  \label{tab:variant-c}
  \small
  \begin{tabular}{lccccc}
    \toprule
    Method & F1@3 & F1@5 & F1@8 & F1@10 & AUC-PR@5 \\
    \midrule
    \textbf{BIDE Coverage (ours)} & \underline{0.661} & \textbf{0.683} & \textbf{0.678} & \textbf{0.678} & 0.729 \\
    Bi-LSTM~\cite{graves2005bilstm} & 0.630 & 0.654 & 0.648 & 0.631 & \textbf{0.741} \\
    Frequency Vector & 0.596 & \underline{0.667} & 0.649 & 0.649 & 0.716 \\
    $n$-gram & 0.562 & 0.568 & 0.571 & 0.571 & \underline{0.736} \\
    TaSPM~\cite{taspm2024} & 0.618 & 0.618 & 0.618 & 0.618 & 0.591 \\
    DeepLog~\cite{du2017deeplog} & 0.430 & 0.500 & 0.535 & 0.529 & 0.610 \\
    Process Conformance & 0.395 & 0.647 & 0.657 & 0.567 & 0.596 \\
    Step Count (control) & \textbf{0.667} & \underline{0.667} & \underline{0.667} & \underline{0.667} & 0.500 \\
    \bottomrule
  \end{tabular}
\end{table}

BIDE produces 33 closed patterns at 5\% minimum support; 32 pass the precision filter (precision range 0.515--1.000). The library matches at least one pattern in 88\% of test-set failures at $K{=}3$. Representative patterns include \texttt{[CLICK\_BID\_SUCCESS\_\_R\_VERIFY]$\times 4$} (validation loop, precision 0.837), \texttt{[TYPE\_BID\_SUCCESS} $\rightarrow$ \texttt{UNKNOWN\_NONE\_SUCCESS} $\rightarrow$ \texttt{TYPE\_BID\_SUCCESS]} (typing-after-parse-failure recovery, precision 0.907), and \texttt{[CLICK\_BID\_SUCCESS]$\times 5$} (navigation without progress, precision 0.808). The four-category taxonomy covers all 32 retained patterns; the closed set is small enough to inspect and organize by hand.

Table~\ref{tab:savings} reports operating points for the online termination policy on the Exclude-errors test set. At the recommended operating point ($K{=}5$, threshold 0.15), the system terminates 50 traces with 92.0\% precision (3.2\% impact on successful traces) and 26.3\% overall token savings. On Variant C at $K{=}3$, threshold 0.10, the same policy achieves perfect precision (15/15) and 12.6\% savings with zero impact on successful tasks. Average savings per terminated trace rise from 50.0\% at threshold 0.05 to 68.0\% at threshold 0.25, since traces terminated earlier consume less of their budget before termination.

\begin{table}[t]
  \centering
  \caption{Online detection operating points on Exclude-errors at $K{=}5$. The boldface row is the recommended operating point: 92\% precision, 3.2\% impact on successful traces, 26.3\% overall token savings.}
  \label{tab:savings}
  \small
  \begin{tabular}{rrrrcccc}
    \toprule
    Threshold & Terminated & TP & FP & Precision & Recall & Success Kill Rate & Overall Savings \\
    \midrule
    0.05 & 129 & 91 & 38 & 0.705 & 0.752 & 0.306 & 54.5\% \\
    0.10 & 103 & 75 & 28 & 0.728 & 0.620 & 0.226 & 46.2\% \\
    \textbf{0.15} & \textbf{50} & \textbf{46} & \textbf{4} & \textbf{0.920} & \textbf{0.380} & \textbf{0.032} & \textbf{26.3\%} \\
    0.20 & 30 & 29 & 1 & 0.967 & 0.240 & 0.008 & 16.2\% \\
    0.25 & 9 & 9 & 0 & 1.000 & 0.074 & 0.000 & 4.4\% \\
    \bottomrule
  \end{tabular}
\end{table}

\section{Limitations}
\label{sec:limitations}

Three limitations are worth noting. First, performance varies substantially across backbones. F1 on the Exclude-errors configuration at $K{=}3$ ranges from 0.526 (Llama-3.2-3B) to 0.649 (Mistral-7B) to 0.913 (Qwen-2.5-7B); at $K{=}10$, the spread is similar (0.480 / 0.629 / 0.913). The library is mined on the pooled training set, so the spread reflects how cleanly each model's failure modes align with population-level patterns rather than per-model overfitting. Qwen's non-success traces are long (mean 14.3 steps) and its successes short (2.84 steps), creating clear behavioral separation; Llama's failures (7.96 steps) and successes (5.02 steps) overlap more, providing less structural signal. The same library cannot be assumed to provide uniform deployment value across backbones.

Second, cross-benchmark transfer is partial and model-dependent. Mean pattern coverage on WebArena shopping traces is 0.100 for Mistral, 0.056 for Llama, and 0.021 for Qwen, inversely correlated with R\_STUCK reasoning-intent frequency (0.8\%, 18.8\%, 42.4\% respectively). Given the three-backbone sample, this may indicate that stronger self-monitoring shifts failures toward reasoning-level signals absent from the action-sequence library. Benchmark-native mining is more appropriate than cross-benchmark transfer.

Third, pattern mining is outperformed by neural sequence models at long $K$ on easy configurations. On Exclude-errors with timeouts retained, Bi-LSTM reaches F1@10 of 0.765, a 15.3 percentage-point gap over BIDE's 0.612. The approaches occupy distinct regions of the deployment trade-off: short $K$ with interpretability requirements favors pattern mining, while long $K$ without diagnostic constraints favors neural classifiers.

Two further caveats apply to the framing of the contribution. The interpretability claim is supported structurally rather than empirically: every termination returns named patterns with positions and a taxonomy category, but we do not report a user study, operator-in-the-loop evaluation, or measurement of whether the repair recommendations improve downstream agent performance. The four-category taxonomy is derived from informal manual inspection of the 32 retained patterns rather than from a documented coding protocol with inter-rater agreement. Detection and savings figures are point estimates from a single 60/20/20 split at a fixed seed; we do not report multi-seed variance or significance tests on the close margins in Table~\ref{tab:variant-c}.

\section{Discussion}
\label{sec:discussion}

Pattern-based detection is most useful in two deployment contexts: short-prefix termination, where a trace cut at step 3 has consumed only $\sim$10\% of a 30-step budget and library compactness makes detection at that prefix length tractable; and operator-facing workflows where the failure category and matched-pattern positions feed downstream tooling such as dashboards, architecture repair, or human-in-the-loop overrides. At the recommended operating point, BIDE clears the deployment thresholds set in advance, with 3.2\% false-positive impact on successful traces (below the 5\% bound) and 26.3\% overall token savings. The multi-variant evaluation framework and step-count baseline are reusable beyond the specific detector reported here, and future early-termination work for web agents should report results on configurations that control for trace-length artifacts.

\begin{ack}
This work was conducted as part of the author's MS thesis at San Jose State University. We thank the SJSU College of Engineering HPC cluster staff for compute support and resources. We thank the AID-Wild reviewers and committee for feedback that improved the camera-ready version of this paper.
\end{ack}

\appendix

\section{Additional Material}
\label{sec:appendix}

This appendix is intended for ``optional reading'' for reviewers. The main paper stands on its own without it.

\subsection{Additional configurations}
The Exclude-errors imbalanced configuration (49.4\% failure prevalence, 1{,}226 traces) shows the same early-detection pattern: BIDE Coverage achieves the highest F1 at $K{=}3$ (0.710) across all eight methods. Bi-LSTM leads at $K{=}10$ (F1 0.758, AUC-PR 0.843); BIDE preserves competitive ranking quality (AUC-PR@10 = 0.766) at the cost of F1.

\subsection{Cross-benchmark coverage}
Mean library-pattern coverage on WebArena shopping traces: Mistral 0.100, Llama 0.056, Qwen 0.021. R\_STUCK reasoning-intent frequency on the same WebArena traces: Qwen 42.4\%, Llama 18.8\%, Mistral 0.8\%, versus 0.2\% on MiniWoB++. The inverse correlation with library coverage suggests R\_STUCK captures a self-monitoring signal that displaces action-sequence patterns mined from MiniWoB++. Pattern-based transfer requires the failure-mode \emph{vocabulary} to overlap, not just the failure-mode \emph{shapes}.

\subsection{Exclude-errors balanced detection}
\label{app:exclude-errors-balanced}

Table~\ref{tab:exclude-errors-balanced} reports the full detection comparison on the Exclude-errors balanced configuration (errors removed; timeouts retained; classes balanced via downsampling). This is the configuration that the abstract's headline F1@$K{=}3$ of 0.717 is drawn from. BIDE Coverage achieves the highest F1 at $K{=}3$ across all eight methods. At longer prefixes, the ranking shifts: Bi-LSTM reaches the highest F1@$K{=}10$ (0.765) and AUC-PR@10 (0.861), at the cost of the interpretability properties that motivate the pattern-mining approach.

\begin{table}[!ht]
  \centering
  \caption{Exclude-errors balanced: F1 at each prefix length and AUC-PR at $K{=}10$. BIDE Coverage leads at $K{=}3$; neural and feature methods overtake at longer prefixes.}
  \label{tab:exclude-errors-balanced}
  \small
  \begin{tabular}{lccccc}
    \toprule
    Method & F1@3 & F1@5 & F1@8 & F1@10 & AUC-PR@10 \\
    \midrule
    \textbf{BIDE Coverage (ours)} & \textbf{0.717} & 0.677 & 0.683 & 0.612 & 0.757 \\
    Frequency Vector & 0.712 & 0.717 & 0.730 & 0.730 & 0.822 \\
    $n$-gram & 0.690 & 0.706 & 0.731 & 0.744 & 0.815 \\
    Bi-LSTM~\cite{graves2005bilstm} & 0.657 & 0.713 & 0.731 & \textbf{0.765} & \textbf{0.861} \\
    TaSPM~\cite{taspm2024} & 0.676 & 0.650 & 0.656 & 0.656 & 0.622 \\
    Step Count (control) & 0.667 & 0.667 & 0.667 & 0.667 & 0.500 \\
    DeepLog~\cite{du2017deeplog} & 0.592 & 0.609 & 0.628 & 0.646 & 0.690 \\
    Process Conformance & 0.626 & 0.654 & 0.452 & 0.472 & 0.581 \\
    \bottomrule
  \end{tabular}
\end{table}

% \newpage
% \input{checklist.tex}

\end{document}